\pdfoutput=1
\documentclass[11pt]{article}

\usepackage{acl}

\usepackage{times}
\usepackage{latexsym}

\usepackage[T1]{fontenc}
\usepackage[utf8]{inputenc}
\usepackage{xcolor}

\usepackage{microtype}
\usepackage{inconsolata}
\usepackage[linesnumbered,ruled,vlined]{algorithm2e}
\usepackage{graphicx}
\usepackage{bm}
\usepackage{enumitem}
\usepackage{multirow}
\usepackage{amsfonts}
\usepackage{amssymb}
\usepackage{amsmath}
\usepackage{verbatim}
\usepackage{booktabs}
\DeclareMathOperator*{\argmax}{arg\,max}

\SetKwInput{KwInput}{Input}                % Set the Input
\SetKwInput{KwOutput}{Output}
\newcommand{\methodname}{\textsc{LitSet}}

\usepackage{cleveref}

\title{Large-Scale Label Interpretation Learning for Few-Shot Named Entity Recognition}

\author{
    Jonas Golde \hspace*{15mm} Felix Hamborg \hspace*{15mm} Alan Akbik \vspace*{2mm}\\
    Humboldt Universität zu Berlin \\
    \texttt{goldejon@informatik.hu-berlin.de} \\
    \texttt{\{felix.hamborg, alan.akbik\}@hu-berlin.de}
}

\begin{document}
\maketitle
\begin{abstract}
Few-shot named entity recognition (NER) detects named entities within text using only a few annotated examples. One promising line of research is to leverage natural language descriptions of each entity type: the common label PER might, for example, be verbalized as ``person entity.'' In an initial \textit{label interpretation learning} phase, the model learns to interpret such verbalized descriptions of entity types. In a subsequent \textit{few-shot tagset extension} phase, this model is then given a description of a previously unseen entity type (such as ``music album'') and optionally a few training examples to perform few-shot NER for this type. In this paper, we systematically explore the impact of a strong semantic prior to interpret verbalizations of new entity types by massively scaling up the number and granularity of entity types used for label interpretation learning. To this end, we leverage an entity linking benchmark to create a dataset with orders of magnitude of more distinct entity types and descriptions as currently used datasets. We find that this increased signal yields strong results in zero- and few-shot NER in in-domain, cross-domain, and even cross-lingual settings. Our findings indicate significant potential for improving few-shot NER through heuristical data-based optimization. 
\end{abstract}

\section{Introduction} \label{sec:introduction}

\begin{figure}[ht]
    \centering
    \includegraphics[width=\linewidth]{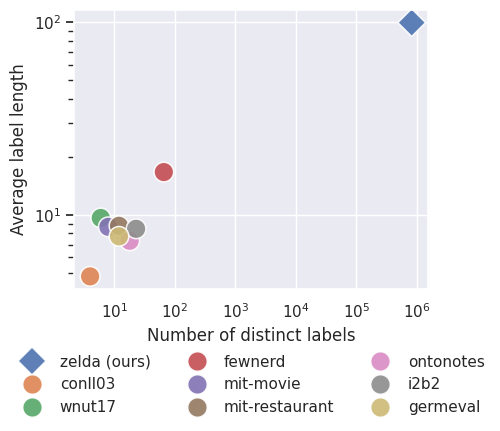}
    \caption{Given existing datasets, few-shot NER methods requiring an initial label interpretation learning are limited regarding entity types and label verbalizations. We propose learning from orders of magnitude more distinct types and more expressive label semantics than current datasets by utilizing ZELDA annotated with WikiData information.}
    \label{fig:overview_datasets}
\end{figure}

Few-shot named entity recognition (NER) refers to identifying and classifying named entities within text by learning from a few annotated examples. A widely adopted strategy in few-shot NER employs transfer learning with pre-trained language models (PLMs) to interpret labels based on their semantic meaning \citep{yang-katiyar-2020-simple,de-lichy-etal-2021-meta,das-etal-2022-container,ma-etal-2022-label,ma-etal-2022-template,ma-etal-2022-decomposed,chen-etal-2023-prompt}. The main idea is that such models learn to interpret a natural language description of an entity type for use in a word-level decoder. They learn in two phases: 

\begin{enumerate}
    \item a \textit{label interpretation learning} phase on a NER-annotated dataset with a set of entity types and their verbalizations. For instance, the common label PER might be verbalized as "person entity." In this phase, the model learns to associate entity type verbalizations with matching NER annotations. 
    \item a \textit{few-shot tagset extension phase} in which the model is expanded to previously unseen domains or entity types using only a new verbalization and optionally a few example annotations. For instance, to extend the model to recognize the names of music albums, one would only need to provide a verbalization ("music album") and a few examples.     
\end{enumerate}

\noindent 
\textbf{Limitations.} However, as Figure~\ref{fig:overview_datasets} indicates, prior studies used only very limited numbers of distinct entity types for label interpretation learning. This is an artifact of relying on common NER datasets such as CoNLL-03 \citep{tjong-kim-sang-de-meulder-2003-introduction}, OntoNotes \citep{pradhan-etal-2012-conll}, WNUT-17 \citep{derczynski-etal-2017-results}, or FewNERD \citep{ding-etal-2021-nerd}, which only contain a small number of distinct entity types (between 4 and 66 types). Furthermore, the majority of their entity types have a simple semantic definition, such as ``person,'' ``location,'' or ``organization,'' and occur across several datasets. We hypothesize that these limitations overly constrain the semantic signal that is observed during label interpretation learning, thus constituting a main limiting factor to few-shot NER.

\noindent 
\textbf{Contributions}. With this paper, we introduce a novel approach named \methodname{} (label interpretation learning by scaling entity types) and systematically investigate the intuition that increasing the number of distinct entity types and their semantic exactness in label interpretation learning introduces a strong semantic prior to understand unseen entities in few-shot settings. To this end, we heuristically create a dataset with orders of magnitude more distinct entity types than commonly employed (cf. ~\Cref{fig:overview_datasets}) and use it for extensive experimentation. In more detail, our contributions are: 

\begin{itemize}
    \item We present experiments to validate our hypothesis on the largest existing NER dataset (FewNERD). We find that few-shot performance increases with label interpretation learning on more distinct entity types and more expressive descriptions (cf.~\Cref{sec:validation_experiment}).
    
    \item We derive a dataset with orders of magnitude more granular entity type annotations to massively scale up label interpretation learning. Our approach leverages the recently released entity linking benchmark ZELDA \citep{milich-akbik-2023-zelda} and enriches it with type descriptions from WikiData \citep{vrandevcic2014wikidata} (cf.~\Cref{sec:method}).
   
    \item We comprehensively evaluate label interpretation learning on our derived corpus against classical setups for zero- and few-shot NER in in-domain, cross-domain, and cross-lingual settings and transfer it to different model architectures (cf.~\Cref{sec:experiments}). 
\end{itemize}

We find that label interpretation learning on our heuristically derived corpus matches and, in many cases, significantly outperforms strong baselines. Our findings indicate significant potential for improving few-shot NER through heuristical data-based optimization. We release the generated dataset and source code under the Apache 2 license on Github\footnote{\href{https://github.com/flairNLP/label-interpretation-learning}{https://github.com/flairNLP/label-interpretation-learning}}.

\section{Validation Experiment for Impact of Entity Types and Label Descriptions} \label{sec:validation_experiment}
%\section{The importance of pre-finetuning in current evaluation setups}

\begin{figure*}[ht]
    \centering
    \includegraphics[width=\textwidth]{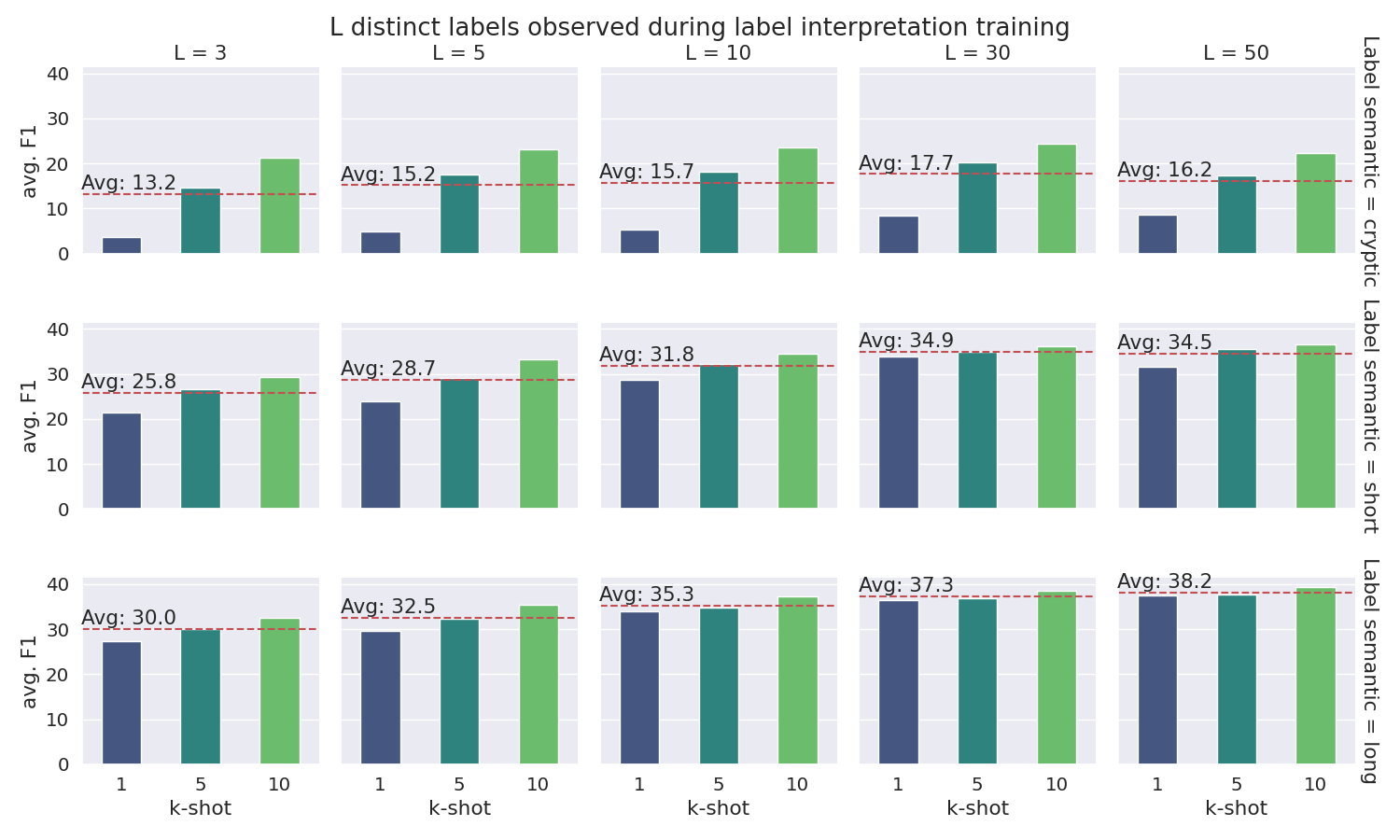}
    \caption{F1 scores for few-shot NER tagset extension on FewNERD depending on how many distinct entity types were seen in label interpretation learning (columns) and how label types were verbalized (rows).  We report F1 scores averaged over five seeds. We observe that (\textit{1}) more distinct labels during label interpretation training and (\textit{2}) more semantically expressive labels improve the few-shot ability on unseen labels.
    %We fix the number of entity mentions to 10k during label interpretation training and employ the identical few-shot splits in all settings.
    }
    \label{fig:motivation_graphic}
\end{figure*}

We first conduct an experiment to validate the intuition that a richer training signal for label interpretation learning positively impacts few-shot NER. 
To this end, we create a set of training datasets for label interpretation learning that each contain the same number of entities but vary in the number of distinct entity types and their label verbalization. We then compare the few-shot NER ability of models trained on each of these datasets.

%label interpretation training impacts the final few-shot performance. To do so, we perform a tag set extension experiment on the FewNERD dataset. We employ the frequently used bi-encoder architecture \citep{blevins-zettlemoyer-2020-moving,ma-etal-2022-label,zhang2023optimizing} as our backbone architecture.
%While previous methods often optimized model architectures, we investigate the impact of the dataset $\mathcal{D}$ with its labels $\mathcal{L}$ used during label interpretation learning on the few-shot performance.

\subsection{Experimental Setup}

\textbf{Definitions.}
To evaluate few-shot NER, an existing dataset $\mathcal{D}$ is split based on its labels $\mathcal{L}$: the label interpretation training split $\mathcal{D}^{LIT}$ and a few-shot fine-tuning split $\mathcal{D}^{FS}$. The corresponding labels of each split $\mathcal{L}^{LIT}$ and $\mathcal{L}^{FS}$ are set such that $\mathcal{L}^{LIT} \cup \mathcal{L}^{FS} = \mathcal{L}$ and $\mathcal{L}^{LIT} \cap \mathcal{L}^{FS} = \emptyset$. 

For few-shot tagset extension, we sample a support set $\mathcal{S}$ by $k$-shot down-sampling $\mathcal{D}^{FS}$. The support set $\mathcal{S}$ contains each label from $\mathcal{L}^{FS}$ exactly $k$ times. We sample three different support sets using different seeds and report the averaged micro-F1 scores over these iterations.

\noindent 
\textbf{Dataset.}
We use FewNERD in our experiment since it is the largest existing dataset w.r.t.~the number of distinct entity types (66 types). We set the labels of $D^{LIT}$ to be the 50 most occurring entity types and the labels of $D^{FS}$ to be the 16 least occurring. We perform an analysis along two dimensions: 

\begin{itemize}
\item To measure the impact of more distinct entity types in label interpretation learning, we create 5 versions of the training data containing 3, 5, 10, 30, and all 50 labels, respectively. Importantly, all versions contain the same number of annotations (10k) to ensure an equal entity detection ability. 
\item To measure the impact of richer verbalizations, we define 3 different labels semantics: \textit{(1)} a "cryptic" unique, random 2-character label, \textit{(2)} a "short" description as regularly used according to research and \textit{(3)} a "long" description with examples (cf. \Cref{sec:label_semantics_for_validation_experiment}).
\end{itemize}

To exclude the respective labels from each split, we follow prior work and mask labels $\mathcal{L}^{LIT}$ in $\mathcal{D}^{FS}$ and $\mathcal{L}^{FS}$ in $\mathcal{D}^{LIT}$ with the \texttt{O}-token (meaning no named entity).

\noindent 
\textbf{Few-shot model.} We employ the frequently used bi-encoder architecture \citep{blevins-zettlemoyer-2020-moving,ma-etal-2022-label} with two \texttt{bert-base-uncased} transformers \citep{NIPS2017_3f5ee243} as our backbone architecture. 

We argue that this architecture has an essential advantage over approaches using cross-attention such as \citet{li-etal-2020-unified,halder-etal-2020-task,chen-etal-2023-prompt}. Previously mentioned methods are limited by the input size of the model (e.g., 512 for BERT) because they prepend label verbalizations to the processed sentence. One could overcome this limitation with one forward pass per label-sentence pair. However, both options become computationally expensive with extensive type descriptions or many distinct entity types. The bi-encoder can be easily adapted to handle an arbitrary number of distinct labels (see ~\Cref{sec:backbone_architecture}).

\begin{figure*}[ht]
    \centering
    \includegraphics[width=\textwidth]{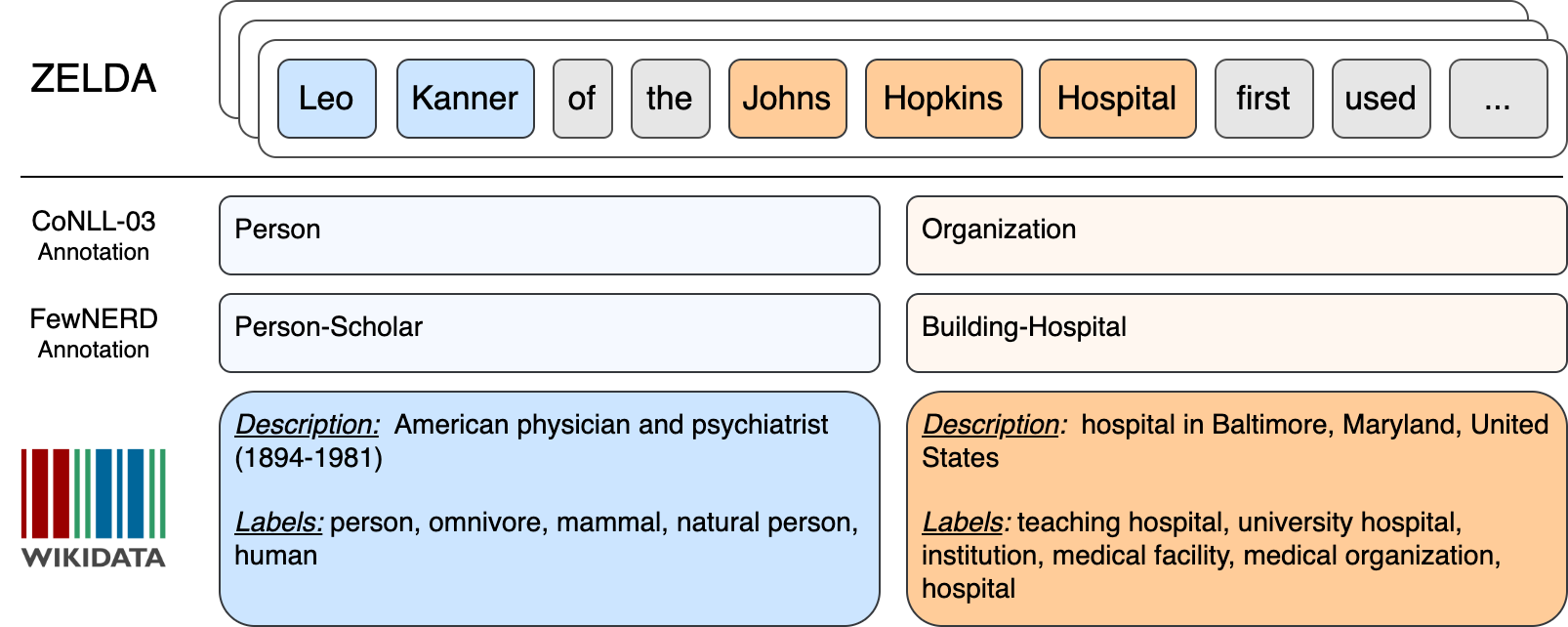}
    \caption{An example annotation of a sentence in ZELDA. WikiData provides precise descriptions and labels about an entity. Annotation types in existing datasets (CoNLL-03, FewNERD) are be less informative if not misleading.}
    \label{fig:zelda_annotation}
\end{figure*}

\subsection{Results}
\Cref{fig:motivation_graphic} shows the results of tagset extension when performing label interpretation learning on FewNERD subsets with different numbers of labels (columns) and different verbalization methods (rows). For each label interpretation learning, we report 
the average F1-score for tagset extension for 1-shot, 5-shot, and 10-shot learning, respectively.  

\noindent 
\textbf{Improved generalization with more types.}
We observe that the number of distinct labels seen during label interpretation training increases the generalization in few-shot settings independent of the label semantics used. We find improvements from +3.0 F1 (cf.~$L$ = 3 vs.~$L$ = 50, label semantic: cryptic) up to +8.7 F1 (cf.~$L$ = 3 vs.~$L$ = 50, label semantic: short) on average in pp. 

\noindent 
\textbf{More expressive descriptions helpful.}
We also find that increasing the expressiveness of label verbalizations strongly improves the few-shot performance. This observation is independent of the distinct number of labels seen in label interpretation learning, such that we find improvements ranging from +16.8 F1 (cf. label semantics: simple vs.~long, with $L$ = 3) up to +22.0 F1 (cf.~label semantics: simple vs.~long, with $L$ = 50) on average in pp. 

These observations on FewNERD confirm our intuition that a richer training signal in label interpretation learning improves few-shot NER performance. To verify this observation for other models, we repeat this experiment with a pre-trained transformer on sparse latent typing, an objective to sparsely extract sentence-level keywords with diverse latent types, where we make the same observation. These experiments are illustrated in detail in~\Cref{sec:appendix_validation_experiment_sparse_latent_typing}.

\section{Large-Scale Label Interpretation Learning} \label{sec:method}

As our validation experiment shows a positive impact of increasing the number and expressivity of entity types, we now aim to scale the signal for label interpretation learning to orders of magnitude more entity types. To this end, we heuristically derive a NER-annotated dataset using the recently released entity linking benchmark ZELDA and annotate it with WikiData information (Section~\ref{sec:underlying_dataset}). We also introduce a modified training procedure for the bi-encoder to handle a very large space of entity types that applies to all architectures of its kind (Section~\ref{sec:backbone_architecture}). We call this approach \methodname{} (label interpretation learning by scaling entity types).

\subsection{\methodname{} Dataset} \label{sec:underlying_dataset}

The task of entity disambiguation is closely related to NER. Here, an already detected entity is disambiguated by linking it to an existing knowledge base such as Wikipedia or WikiData. Existing training and evaluation datasets for entity disambiguation thus contain named entities marked with links to entries in the WikiData knowledge base. 

One advantage of WikiData is that it contains fine-grained labels and free-form text descriptions of entities in the knowledge base. For instance, the entity "John Hopkins Hospital" (cf.~\Cref{fig:zelda_annotation}) has the free-form description "hospital in Baltimore, Maryland" and belongs to the classes "teaching hospital", "university hospital", and many others. As the Figure shows, these labels are significantly more fine-grained than CoNLL-03 and even FewNERD entity types which simply classify it as an "organization" or a "hospital" respectively. 

\begin{table}
\small
\centering
\begin{tabular}{lcc}
\toprule
\textbf{Dataset} & \textbf{Label length} & \textbf{\# Distinct types}\\
\midrule
CoNLL-03 & $9.8 \pm 2.9$ & 4 \\
WNUT17 & $8.3 \pm 2.8$ & 6 \\
OntoNotes & $9.8 \pm 8.5$ & 18 \\
FewNERD & $17.3 \pm 7.6$ & 66 \\
\midrule
\methodname{} & $99.8 \pm 45.4$ & \textasciitilde 817k \\
\bottomrule
\end{tabular}
\caption{Average label description length (in characters) and distinct entity types of NER datasets. Label length and distinct entity types for \methodname{} refers to all annotations as indicated in~\Cref{fig:zelda_annotation}.}
\label{tab:average_description_length}
\end{table}

\noindent 
\textbf{Deriving the dataset.} We leverage the classes and descriptions from WikiData as type annotations in our approach. For each linked entity in the dataset, we retrieve the types and descriptions from WikiData  and use them as NER annotations. We refer to~\Cref{sec:appendix_prefinetuning_labels} for a detailed explanation of the fields used.

To best prepare our model for arbitrary labels in a few-shot setting, we sample the annotations to learn to interpret annotations on different hierarchies. We assume labels to represent high-level types, whereas descriptions are very specific to that entity. Specifically, for each entity $x_i$, we uniformly sample whether we annotate it with either the description attribute or the labels attribute (cf.~\Cref{fig:zelda_annotation}). When utilizing the labels attribute, we randomly select the number of tags following a geometric distribution with $p = .5$. Subsequently, we uniformly sample tags from the label attribute until the number of tags is reached. Lastly, we concatenate the selected tags for final annotation.

\subsection{Backbone Architecture} \label{sec:backbone_architecture}
Due to its simplicity, we conduct our experiments using the widely adopted bi-encoder model. It utilizes two separate transformers to encode tokens and labels, respectively. The first transformer generates embeddings $e_t \in \mathbb{R}^{N \times H}$ for all tokens, where $N$ represents the number of tokens and $H$ denotes the hidden size of the model. The second obtains the \texttt{[CLS]}-token embeddings $e_l$ for the labels converted into natural language. We employ cross-entropy loss and derive final predictions with

$$\hat{y} = \argmax softmax(e_t \cdot e_l)$$ 

However, training a model, including the bi-encoder, with a wide array of distinct classes is non-trivial. With $\mathcal{L}$ denoting the set of labels, the shape of label representations is $e_l \in \mathbb{R}^{|\mathcal{L}|\times H}$. Given that $|\mathcal{L}| \approx 10^{6}$ (cf. \Cref{fig:overview_datasets}), we aim to circumvent the resulting matrix multiplication for two reasons: (\textit{1}) computational limitations and (\textit{2}) optimization difficulty. To alleviate these issues, we restrict our consideration to labels present in the current batch $\mathcal{L}_b$ with $|\mathcal{L}_b| \ll |\mathcal{L}|$ for loss calculation.

\section{Experiments} \label{sec:experiments}

We evaluate the impact of label interpretation training in various tagset extension settings. Throughout all experiments, we compare label interpretation learning on \methodname{} with training on different baseline datasets. We present all hyperparameters used for our experiments in~\Cref{sec:lit_hyperparameters}. Specifically, we conduct the following experiments:

\begin{enumerate}

\item \textit{In-domain transfer}: Identical domain in label interpretation learning and few-shot fine-tuning (cf.~\Cref{sec:extension_in_domain}).

\item \textit{Cross-domain transfer}: Different domain in label interpretation learning and few-shot fine-tuning (cf. \Cref{sec:extension_out_domain}).

\item \textit{Transfer to advanced bi-encoders}: Identical to in-domain setting, but we transfer our approach to advanced bi-encoder architectures (cf. \Cref{sec:advanced_architectures}).

\item \textit{Cross-lingual transfer}: Identical domain in label interpretation learning and few-shot fine-tuning, but languages differ between both phases (cf. \Cref{sec:cross_lingual}).

\end{enumerate}

Further, we support our experiments by analyzing the impact of different label semantics used between label interpretation learning and few-shot fine-tuning (cf.~\Cref{sec:extension_in_domain}). At last, we refer to our ablation experiments using (\textit{1}) different transformers as label encoders and (\textit{2}) negative sampling (cf. \Cref{sec:ablation_sbert,sec:ablation_negative_sampling}).

\begin{figure}[t!]
    \centering
    \includegraphics[width=\linewidth]{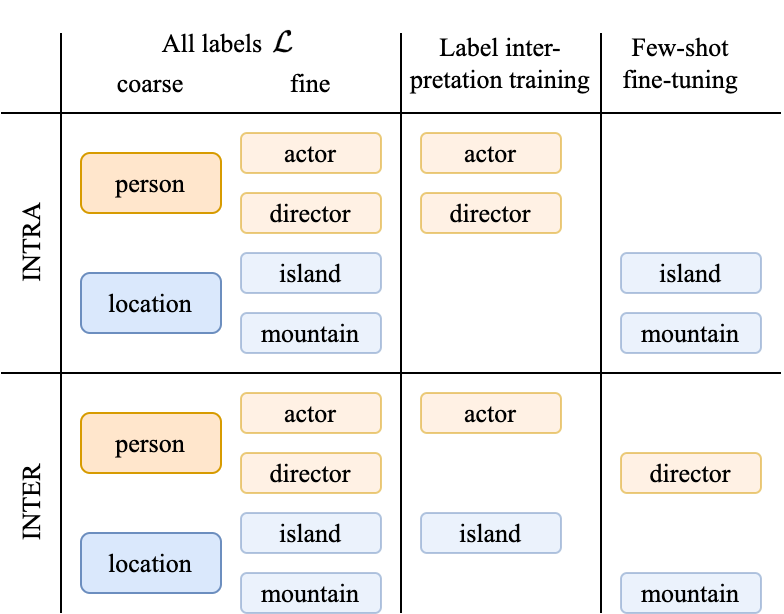}
    \caption{Exemplary illustration on the INTRA and INTER settings of FewNERD experiments.}
    \label{fig:inta_inter_setting}
\end{figure}

\subsection{Experiment 1: In-Domain Transfer} \label{sec:extension_in_domain}

This experiment replicates the most common evaluation setup for few-shot tagset extension, where both $\mathcal{D}^{LIT}$ and $\mathcal{D}^{FS}$ are sourced from the same NER dataset. Our baseline is the default approach of label interpretation learning on $\mathcal{D}^{LIT}$, which is "in-domain" since it shares the same textual domain and entity annotations are aligned on identical semantic levels as the evaluation data, whereas label interpretation learning on \methodname{} does not have these advantages.

\subsubsection{Experimental Setup} 

\begin{table*}[ht]
\centering
\small
\renewcommand{\arraystretch}{1.2}
\setlength{\tabcolsep}{5pt}
\begin{tabular}{p{3.1cm}p{3.7cm}ccccr}
\toprule
Evaluation data $\mathcal{D}^{FS}$ for tagset extension from:  & Label interpretation learning data $\mathcal{D}^{LIT}$ from: & 0-shot & 1-shot & 5-shot & 10-shot & Avg. \\
\toprule
\multirow{5}{*}{OntoNotes} & \methodname{}  & $\bm{8.7} \pm 1.7$ & $\bm{21.9} \pm 8.4$ & $\bm{40.1} \pm 7.2$ & $\underline{48.4} \pm 6.2$ & $\bm{29.5}$ \\
& \hspace{0.3cm} w/ all labels & $3.5 \pm 1.3$ & $\underline{20.0} \pm 9.5$ & $\underline{38.4} \pm 8.3$ & $46.5 \pm 6.3$ & $\underline{27.1}$ \\
& \hspace{0.3cm} w/ labels only  & $0.1 \pm 0.1$ & $14.3 \pm 8.3$ & $29.6 \pm 6.9$ & $37.5 \pm 6.1$ & $20.4$ \\
& \hspace{0.3cm} w/ description only & $\underline{4.2} \pm 1.3$ & $19.8 \pm 8.8$ & $37.5 \pm 7.9$ & $46.2 \pm 5.9$ & $26.9$ \\
& OntoNotes \textit{(Baseline)} & $0.2 \pm 0.1$ & $11.2 \pm 9.3$ & $38.3 \pm 12.0$ & $\bm{54.9} \pm 7.6$ & $26.2$ \\
\midrule
\multirow{5}{*}{FewNERD$_{\textsc{Intra}}$}  & \methodname{} & $3.2 \pm 1.0$ & $\bm{30.7} \pm 5.3$ & $\bm{51.9} \pm 5.2$ & $\bm{57.9} \pm 6.2$ & $\bm{35.9}$ \\
& \hspace{0.3cm} w/ all labels & $0.9 \pm 0.4$ & $\underline{20.1} \pm 5.0$ & $\underline{47.7} \pm 6.0$ & $\underline{54.1} \pm 5.9$ & $\underline{30.7}$ \\
& \hspace{0.3cm} w/ labels only & $\underline{3.7} \pm 0.5$ & $14.3 \pm 8.3$ & $29.6 \pm 7.0$ & $37.5 \pm 6.1$ & $21.3$ \\
& \hspace{0.3cm} w/ description only & $1.0 \pm 0.3$ & $19.8 \pm 8.8$ & $37.5 \pm 7.9$ & $46.2 \pm 5.9$ & $26.1$ \\ 
& FewNERD$_{\textsc{Intra}}$ \textit{(Baseline)} & $\bm{5.8} \pm 0.4$ & $8.9 \pm 4.3$ & $31.4 \pm 9.2$ & $38.4 \pm 7.5$ & $21.1$ \\ 
\midrule
\multirow{5}{*}{FewNERD$_{\textsc{Inter}}$}  & \methodname{} & $\bm{24.3} \pm 0.6$ & $\bm{39.8} \pm 2.9$ & $\underline{49.1} \pm 1.9$ & $\underline{52.1} \pm 1.9$ & $\bm{41.3}$ \\
& \hspace{0.3cm} w/ all labels & $\underline{17.6} \pm 2.5$ & $36.1 \pm 4.7$ & $47.2 \pm 3.0$ & $50.4 \pm 2.4$ & $37.8$ \\
& \hspace{0.3cm} w/ labels only & $2.9 \pm 0.6$ & $24.7 \pm 1.8$ & $37.9 \pm 1.7$ & $42.4 \pm 2.0$ & $27.2$ \\
& \hspace{0.3cm} w/ description only & $16.2 \pm 2.0$ & $37.4 \pm 2.9$ & $47.8 \pm 2.2$ & $50.9 \pm 1.9$ & $38.1$ \\
& FewNERD$_{\textsc{Inter}}$ \textit{(Baseline)} & $10.6 \pm 0.8$ & $\underline{38.4} \pm 3.1$ & $\bm{50.4} \pm 3.1$ & $\bm{53.3} \pm 2.6$ & $\underline{38.2}$ \\
\bottomrule
\end{tabular}
\caption{Evaluation of zero- and few-shot tagset extension for in-domain settings. We compare the baseline approach of using in-domain data for label interpretation learning against using \methodname{}. Despite lacking the in-domain advantage of the baselines, training on \methodname{}  matches or significantly outperforms the in-domain baseline in nearly all settings. Best scores are in bold, and 2nd best is underlined.}\label{table:extension_in_domain}
\end{table*}

We use OntoNotes and FewNERD in our experiments as they have important properties: OntoNotes covers multiple domains and languages such that we can measure the transferability of our approach. FewNERD comes with two annotation layers: coarse labels $\mathcal{L}^c$ (8 classes) and fine labels $\mathcal{L}^f$ (66 classes). $\mathcal{L}^f$ are subclasses of the $\mathcal{L}^c$ such that the entity mentions of both annotations are identical, only their surface form differs. Thus, we can evaluate our dataset against FewNERD in two ways: \textit{(1)} in the INTRA setting in which we split the labels based on coarse annotations, and \textit{(2)} in the INTER setting in which we split based on the fine annotations (cf. \Cref{fig:inta_inter_setting}). 

We split each dataset into two equally sized label sets for both settings. The random split of labels is repeated three times to reduce the impact of randomness. We then perform few-shot fine-tuning runs with three different seeds for each random split.

\noindent 
\textbf{Comparison with \methodname{}.} To focus solely on understanding the impact of scaling entity types without the influence of increased entity detection, we downsample \methodname{} to match the number of entity mentions in each baseline dataset. Further, to make a fair comparison, we remove labels from our approach that match those in the baseline labels $\mathcal{L}^{FS}$ and mask them with the \texttt{O}-token. However, due to our sampling method, \methodname{} annotations may not always be consistent. Thus, we can only ensure excluding exact overlaps with the few-shot domain.

\subsubsection{Results}
The experimental results are shown in \Cref{table:extension_in_domain}, and we find that \methodname{} substantially improves the few-shot performance in in-domain settings. 

\noindent
\textbf{Detecting coarse entity types.}
 When performing label interpretation learning on OntoNotes and FewNERD$_{\textsc{Intra}}$, we evaluate the model's ability to identify entirely new concepts (see INTRA in~\Cref{fig:inta_inter_setting}). The results in~\Cref{table:extension_in_domain} show that our approach can effectively leverage its general label interpretation ability to outperform baselines by large margins. We report +14.8 F1 on average in .pp on FewNERD$_{\textsc{Intra}}$ and +3.3 F1 on OntoNotes. While \methodname{} consistently outperforms in-domain label interpretation learning on FewNERD (INTRA), this advantage levels off when k = 10 on OntoNotes.

\noindent
\textbf{Differentiating fine entity types.}
 In this setting, the model is exposed to sub-classes of a coarse category during label interpretation learning (e.g., ``actor'' is a subclass of ``person'', cf. INTER in~\Cref{fig:inta_inter_setting}). We observe that all approaches yield improved few-shot generalization in this setting. This finding suggests that transfer to unseen labels is particularly effective when the training includes annotations of high-level categories. With \methodname{}, we outperform FewNERD$_{\textsc{Inter}}$ in 0- and 1-shot settings (+13.7 F1 and +1.4 F1 on average in pp.) and remain competitive at higher k-shots.
 
\begin{table*}[ht]
\centering
\small
\renewcommand{\arraystretch}{1.2}
\setlength{\tabcolsep}{5pt}
\begin{tabular}{p{3.1cm}p{3.7cm}ccccr}
\toprule
Evaluation data $\mathcal{D}^{FS}$ for tagset extension from:  & Label interpretation learning data $\mathcal{D}^{LIT}$ from: & 0-shot & 1-shot & 5-shot & 10-shot & Avg. \\
\toprule
\multirow{3}{*}{JNLPBA} & \methodname{} & $\underline{41.3} \pm 2.0$ & $\underline{25.4} \pm 5.3$ & $\bm{51.3} \pm 3.4$ & $\bm{57.7} \pm 3.0$ & $\bm{43.9}$  \\
& \hspace{0.3cm} w/ all labels & $\bm{42.2} \pm 1.8$ & $22.5 \pm 8.1$ & $\underline{49.9} \pm 3.8$ & $\underline{55.8} \pm 2.7$ & $\underline{42.6}$  \\
& FewNERD$_{\textsc{Inter}}$ & $8.2 \pm 1.5$ & $\bm{29.5} \pm 15.0$ & $46.0 \pm 7.6$ & $49.7 \pm 6.6$ & $33.4$ \\
\midrule
%\multirow{3}{*}{CLUB} & FewNERD$_{\textsc{Inter}}$ & $1.7 \pm 0.2$ & $16.9 \pm 1.8$ & $25.5 \pm 4.9$ & $32.2 \pm 3.7$ & $19.1$ \\ 
\multirow{3}{*}{CLUB} & \methodname{} & $\underline{6.1} \pm 0.9$ & $\underline{19.4} \pm 3.3$ & $\underline{25.9} \pm 3.7$ & $\underline{33.0} \pm 2.1$ & $\underline{21.1}$  \\
& \hspace{0.3cm} w/ all labels & $\bm{7.3} \pm 0.1$ & $\bm{19.9} \pm 2.0$ & $\bm{27.6} \pm 4.6$ & $\bm{35.1} \pm 3.1$ & $\bm{22.5}$  \\
 & FewNERD$_{\textsc{Inter}}$ & $1.7 \pm 0.2$ & $16.9 \pm 1.8$ & $25.5 \pm 4.9$ & $32.2 \pm 3.7$ & $19.1$ \\ 
\bottomrule
\end{tabular}
\caption{\methodname{} outperforms FewNERD in out-of-domain settings on JNLPBA (bio-medical domain) and CLUB (chemical domain).}
\label{table:extension_out_domain}
\end{table*}

\noindent
\textbf{Impact of \methodname{} sampling.}
We measure the impact of different heuristics for creating \methodname{} types. To test this, we conduct various experiments using \methodname{} with (\textit{1}) only labels, (\textit{2}) only descriptions, and (\textit{3}) all label information available (cf.~\Cref{fig:zelda_annotation}). We first find that using only label annotations decreases performance compared to the baselines (cf.~FewNERD$_{\textsc{Inter}}$ and OntoNotes), underlining the need for precise label semantics during label interpretation training to obtain a strong few-shot generalization. 

When using only the descriptions or all available annotations, we notice that \methodname{} yields similar performance to their respective baselines, whereas in the FewNERD$_{\textsc{Intra}}$ setting, substantial improvements are observed compared to the baselines. Again, this emphasizes that learning from detailed label semantics before the few-shot transfer improves the final performance. 

At last, we observe that \methodname{} substantially outperforms all baselines using our sampling technique, which indicates that alternating shorter labels and expressive short descriptions achieves the best generalization.

\subsection{Experiment 2: Cross-Domain Transfer} \label{sec:extension_out_domain}

This experiment assesses the performance of \methodname{} and its corresponding baselines when not only tagsets but also domains of label interpretation learning and few-shot fine-tuning differ. We re-use \methodname{} and FewNERD$_{\textsc{Inter}}$ models after label interpretation learning from previous experiment and evaluate on out-of-domain datasets JNLPBA \citep{collier-kim-2004-introduction} (bio-medical domain) and the Chemical Language Understanding Benchmark (CLUB) \citep{kim-etal-2023-chemical} (chemical domain) which labels do represent entirely new, domain-specific concepts.

\begin{table*}[ht]
\centering
\small
\renewcommand{\arraystretch}{1.2}
\begin{tabular}{lp{2.3cm}p{2.6cm}cccr}
\toprule
Model & Tagset extension on $\mathcal{D}^{FS}$ & Label interpretation learning on $\mathcal{D}^{LIT}$ & 1-shot & 5-shot & 10-shot & Avg. \\
\toprule
\multirow{4}{*}{LEAR} & \multirow{2}{*}{FewNERD$_{\textsc{Intra}}$}  & \methodname{} & $\bm{16.6} \pm 4.2$ & $\bm{33.2} \pm 9.2$ & $\bm{43.4} \pm 10.8$ & $\bm{31.1}$ \\
& & FewNERD$_{\textsc{Intra}}$ & $13.5 \pm 9.2$ & $23.7 \pm 11.7$ & $37.0 \pm 14.6$ & $24.7$ \\ 
\cline{2-7}
& \multirow{2}{*}{FewNERD$_{\textsc{Inter}}$}  & \methodname{} & $14.1 \pm 2.2$ & $38.3 \pm 3.3$ & $44.1 \pm 2.6$ & $32.2$ \\
& & FewNERD$_{\textsc{Inter}}$ & $\bm{27.6} \pm 4.6$ & $\bm{50.8} \pm 3.5$ & $\bm{54.8} \pm 2.6$ & $\bm{44.4}$ \\
\midrule
\multirow{4}{*}{BINDER} & \multirow{2}{*}{FewNERD$_{\textsc{Intra}}$}  & \methodname{} & $\bm{18.8} \pm 6.2$ & $\bm{31.0} \pm 4.2$ & $\bm{33.8} \pm 3.7$ & $\bm{27.9}$ \\
& & FewNERD$_{\textsc{Intra}}$ & $2.6 \pm 1.3$ & $11.5 \pm 5.6$ & $20.7 \pm 7.0$ & $11.6$ \\ 
\cline{2-7}
& \multirow{2}{*}{FewNERD$_{\textsc{Inter}}$}  & \methodname{} & $\bm{18.6} \pm 1.5$ & $\bm{27.3} \pm 1.8$ & $\bm{30.4} \pm 2.0$ & $\bm{25.4}$ \\
& & FewNERD$_{\textsc{Inter}}$ & $6.1 \pm 0.9$ & $20.2 \pm 3.2$ & $26.6 \pm 3.4$ & $17.6$ \\
\bottomrule
\end{tabular}
\caption{Transfer of \methodname{} to advanced bi-encoder architectures. We outperform baselines when coarse entity types are not learned during label interpretation training. On BINDER, we also improve over in-domain label interpretation learning.}\label{table:adaoption_to_other_architectures}
\end{table*}

\subsubsection{Results}
\Cref{table:extension_out_domain} shows the results for cross-domain settings. While this setting is identical for \methodname{}, the baseline now has no advantage of exposure to "in-domain" data during label interpretation training. Further, no additional masking is required since label spaces between JNLPBA and the baseline model are disjoint. Consequently, we do not mask any labels in \methodname{} to maintain a fair comparison. However, we emphasize that our model may have been exposed to close domain-specific labels during label interpretation training.

\noindent
\textbf{\methodname{} better transfers to new domains.}
We find that \methodname{} significantly outperforms FewNERD with average improvements of +10.5 F1 on JNLPBA and +3.4 F1 on CLUB. Further, on JNLPBA, we observe that our sampling approach performs slightly better than using all label information, whereas we observe the opposite when evaluating CLUB. Our approach consistently outperforms FewNERD on CLUB and JNLPBA with higher shots ($k$ >= 5) and achieves an average increase of +34.0 F1 pp. in zero-shot settings on JNLPBA.

\noindent
\textbf{Impact of inconsistent annotations.}
Furthermore, we observe that \methodname{} underperforms by -4.1 F1 pp. compared to the baseline in 1-shot settings on JNLPBA. Additionally, its performance is inferior even compared to the 0-shot scenario. This indicates the instability of few-shot fine-tuning with \methodname{} at very low $k$. Upon further qualitative analysis of the generated dataset, we discovered that annotations from entity linking benchmarks like ZELDA may not be consistently annotated (cf. \Cref{sec:annotation_noise}). This inconsistency could be one possible reason for the observed performance drops. However, as $k$ increases, our approach demonstrates the ability to adapt to the target domain.

\subsection{Experiment 3: Transfer to Advanced Bi-Encoders}\label{sec:advanced_architectures}

This experiment extends our approach to advanced bi-encoder architectures LEAR \citep{yang-etal-2021-enhanced} and BINDER \citep{zhang2023optimizing}. Instead of matrix multiplication, LEAR implements a self-attention layer between the token and label encoder, whereas BINDER uses a contrastive loss. The experimental setup is equal to the one from~\Cref{sec:extension_in_domain}.

\subsubsection{Results}
The results are shown in~\Cref{table:adaoption_to_other_architectures}. We find that \methodname{} with LEAR improves over the corresponding baseline in INTRA settings up to +9.5 F1 on average in pp. at $k$ = 5. Notably, both the baseline and our approach exhibit relatively diminished performance compared to results in~\Cref{sec:extension_in_domain}. However, our approach falls short in INTER settings, confirming our earlier experimental findings. A noteworthy enhancement is discerned at $k$=10 for the baseline in the INTER-setting, suggesting that existing architectures excel in in-domain transfer, particularly when labels closely align. However, in more practical settings (cross-domain and entirely new type concepts), \methodname{} works well with LEAR.

Further, we surpass baselines in INTRA and INTER settings across all $k$-shots for BINDER, indicating \methodname{} also applies to metric-based methods using contrastive objectives. However, to the best of our knowledge, we are the first to evaluate BINDER in such transfer settings. Our evaluation reveals that the overall performance lags behind simpler architectures. We note that BINDER's contrastive loss is tailored for learning from extensively annotated corpora. Thus, BINDER may require modifications or extensions for good generalization performance in these transfer scenarios.

\subsection{Experiment 4: Cross-Lingual Transfer} \label{sec:cross_lingual}

In this experiment, we utilize the multilingual \texttt{xlm-roberta-base} model \citep{conneau-etal-2020-unsupervised} to assess the transferability of \methodname{} across languages. We use the English version of OntoNotes as the baseline for label interpretation training. ZELDA is also an English corpus. The transfer is done on the Arabic and Chinese versions of OntoNotes. The results are shown in Table \ref{table:cross_lingual}. 

\subsubsection{Results}
We find strong improvements across all $k$-shots on the Arabic and Chinese segments of OntoNotes, namely +3.9 F1 and +9.0 F1 on average in pp., respectively. Despite the overlapping domains between label interpretation learning and few-shot fine-tuning on OntoNotes, our model can discern subtle annotation differences across languages. This emphasizes our model's robust understanding of labels in multilingual scenarios.

Furthermore, we observe that utilizing \texttt{xlm-roberta-base} also improves \methodname{}'s performance in monolingual settings (cf.~ \Cref{sec:extension_in_domain}). We reduce the previous performance gap at $k$ = 10 from -6.5 F1 to -0.5 F1 on average in pp., thereby increasing the overall performance from +3.3 F1 to +6.5 F1.

\begin{table*}[ht]
\centering
\small
\renewcommand{\arraystretch}{1.2}
\setlength{\tabcolsep}{5pt}
\begin{tabular}{p{3.1cm}p{3.7cm}ccccr}
\toprule
Evaluation data $\mathcal{D}^{FS}$ for tagset extension from:  & Label interpretation learning data $\mathcal{D}^{LIT}$ from: & 0-shot & 1-shot & 5-shot & 10-shot & Avg. \\
\toprule
\multirow{2}{*}{OntoNotes (EN)} & \methodname{} (EN) & $\bm{9.9} \pm 3.2$ & $\bm{27.4} \pm 8.5$ & $\bm{46.4} \pm 6.7$ & $55.5 \pm 6.4$ & $\bm{34.8}$ \\
 & OntoNotes (EN) & $0.3 \pm 0.1$ & $15.9 \pm 8.4$ & $41.1 \pm 15.0$ & $\bm{56.0} \pm 12.7$ & $28.3$ \\
\midrule
\multirow{2}{*}{Ontonotes (AR)} & \methodname{} (EN) & $0.0 \pm 0.0$ & $\bm{7.2} \pm 6.1$ & $\bm{14.8} \pm 6.3$ & $\bm{22.0} \pm 5.8$ & $\bm{14.7}$ \\
 & Ontonotes (EN) & $0.0 \pm 0.0$ & $4.7 \pm 4.7$ & $12.8 \pm 4.8$ & $14.9 \pm 7.9$ & $10.8$ \\
\midrule
\multirow{2}{*}{Ontonotes (ZH)} & \methodname{} (EN) & $\bm{3.0} \pm 0.9$ & $\bm{22.7} \pm 8.6$ & $\bm{37.6} \pm 5.0$ & $\bm{42.8} \pm 5.0$ & $\bm{26.5}$ \\
 & Ontonotes (EN) & $1.6 \pm 0.3$ & $10.8 \pm 5.9$ & $26.2 \pm 6.9$ & $31.2 \pm 7.9$ & $17.5$ \\
\bottomrule
\end{tabular}
\caption{Tag set extension with baseline pre-finetuning and few-shot fine-tuning in the same domain. \methodname{} outperforms models that are pre-finetuning on in-domain data when pre-finetuning is done on a small number of labels.}
\label{table:cross_lingual}
\end{table*}

\section{Related Work}
Despite advancements achieved through pre-trained word embeddings \citep{peters-etal-2018-deep,akbik-etal-2018-contextual,devlin-etal-2019-bert,liu2019roberta,yamada-etal-2020-luke,raffel2020exploring}, few-shot NER focuses explicitly on generalizing to previously unseen label categories by leveraging a small number of labeled examples.

Metric learning \citep{vinyals2016oneshot,snell2017proto} is a common approach for few-shot NER \citep{fritzerl2019protoNER,wiseman-stratos-2019-label,ziyadi2020examplebased} and employs a distance metric to learn a shared representation space and assign labels based on class prototypes \citep{yang-katiyar-2020-simple,hou-etal-2020-shot,ma-etal-2022-label,han2023metalearning}. Additional components like contrastive loss \citep{das-etal-2022-container,layegh2023contrastner} or meta-learning \citep{de-lichy-etal-2021-meta,ma-etal-2022-decomposed,wang-etal-2022-enhanced} often further improve the performance. Our approach aligns with this research by employing the bi-encoder architecture proposed in \citet{ma-etal-2022-label} with an adapted loss calculation. However, prior work did not investigate the impact of the dataset used for label interpretation learning. We do so by increasing the training signal with expressive label verbalizations. Thus, our approach may be applied to all prior work that relies on label verbalizations but may require architectural adaptations to accommodate arbitrary labels.

Template-filling and prompting methods with (large) language models \citep{lewis-etal-2020-bart,brown2020language,raffel2020exploring,workshop2023bloom,touvron2023llama} have been widely used for few-shot NER \citep{cui-etal-2021-template,ma-etal-2022-template,lee-etal-2022-good,kondragunta-etal-2023-improving,ma2023large}. However, these approaches, relying on masked language model (MLM) objectives, may not be directly comparable to our method due to the scale of our labels. In its basic form, the template-based approach requires one forward pass per label or is limited by the model's maximum sequence length. Additionally, our approach does not depend on large language models, which are often unavailable or impractical for few-shot NER.

While specific efforts have been made to adapt to tags in few-shot domains \citep{hu-etal-2022-label,ji-etal-2022-shot}, these studies evaluated only a limited number of labels. Our approach shares similarities with \citep{ren-etal-2022-language} and \citet{chen-etal-2022-shot}, where models were pre-trained using event mentions and entity links, respectively. However, our approach differs significantly. In \citet{ren-etal-2022-language}, the pre-training objective targets the latent typing of entities, whereas our approach focuses on explicitly scaling up entity typing of few-shot NER models. Our distinction from \citet{chen-etal-2022-shot} lies in exploring the effectiveness of distantly supervised training in a genuine few-shot context, wherein classes are not observed during label interpretation training.

\section{Conclusion}

This paper introduces \methodname{}, a novel approach for label interpretation training with a large-scale set of entity types. We utilize an entity linking dataset annotated with WikiData information, resulting in a  dataset with significantly more distinct labels. We conducted a thorough heuristical, data-based optimization of few-shot NER models using \methodname{}. Our experiments demonstrate that \methodname{} consistently outperforms various in-domain, cross-domain, and cross-lingual baselines and is transferable to other architectures and transformer models. For example, we surpass FewNERD by +14.7 F1 on average in pp. and Chinese OntoNotes by +9.0 F1 on average in pp. in low-resource settings. Our method and experiments provide valuable insights into the factors influencing the performance of few-shot NER models utilizing label semantics.

\clearpage

\section*{Limitations}
Our heuristic data-based optimization is an initial exploration of the impact of scaling the number of distinct entity types during label interpretation learning on few-shot capability. Given our focus on this optimization, we select a commonly used backbone architecture and one entity linking dataset. While we achieved substantial improvements in many settings, it is noteworthy that we did not explore all entity linking benchmarks. Thus, applying our approach with different model architectures and entity disambiguation datasets may yield significantly varied results. Further investigation is necessary to understand how these factors interact comprehensively and to develop more generalized few-shot NER models and comparable evaluation settings.

Additionally, achieving 0-shot capability on completely unseen tags remains challenging, especially in languages different from the one used for label interpretation training. This limitation highlights the need for future research and exploring innovative techniques to enhance the adaptability of few-shot NER models in 0-shot scenarios, enabling them to handle diverse domains and situations effectively.

Lastly, concerning \methodname{}, our best results were obtained by learning solely from in-batch instances. Although this strategy is commonly employed in machine learning, there is substantial related work on learning from negatives, such as contrastive learning. We believe exploring other architectures and loss functions in more detail, including those from contrastive learning, could further improve our method.

\section*{Ethics Statement}
In our opinion, this work does not raise many ethical problems. One primary concern is that the texts of entity linking datasets serving our approach show signs of bias. If not checked correctly in advance, the model may learn these biases as exemplarily shown in \citet{haller2023opiniongpt}. 

\section*{Acknowledgements}
We thank all reviewers for their valuable comments. Jonas Golde is supported by the German Federal Ministry of Economic Affairs and Climate Action (BMWK) as part of the project ENA (KK5148001LB0). Felix Hamborg is supported by the WIN program of the Heidelberg Academy of Sciences and Humanities, financed by the Ministry of Science, Research and Arts of the State of Baden-Wurttemberg, Germany. Alan Akbik is supported by the Deutsche Forschungsgemeinschaft (DFG, German Research Foundation) under Emmy Noether grant ``Eidetic Representations of Natural Language'' (project number 448414230) and under Germany’s Excellence Strategy "Science of Intelligence" (EXC 2002/1, project number 390523135).

\bibliography{anthology}

\newpage
\appendix

\section*{Appendix}

\section{FewNERD Label Semantics in Validation Experiment} \label{sec:label_semantics_for_validation_experiment}

\Cref{table:simple_fewnerd_labels,table:standard_fewnerd_labels,table:long_fewnerd_labels} show an overview of the label semantics used in our validation experiment.

\begin{table}[ht]
\centering
\small
\renewcommand{\arraystretch}{1.15}
\begin{tabular}{lp{3cm}}
Original Label & Adapted Label \\
\toprule
O & XO \\
location-GPE & PH \\
person-politician & EX \\
organization-education & CE \\
\bottomrule
\end{tabular}
\caption{Extract of random two letter labels for FewNERD.}
\label{table:simple_fewnerd_labels}
\end{table}

\begin{table}[ht]
\centering
\small
\renewcommand{\arraystretch}{1.15}
\begin{tabular}{lp{3cm}}
Original Label & Adapted Label \\
\toprule
O & XO \\
location-GPE & geographical social-political entity \\
person-politician & politician \\
organization-education & education \\
\bottomrule
\end{tabular}
\caption{Extract of short labels for FewNERD.}
\label{table:standard_fewnerd_labels}
\end{table}

\begin{table}[ht]
\centering
\small
\renewcommand{\arraystretch}{1.15}
\begin{tabular}{lp{3cm}}
Original Label & Adapted Label \\
\toprule
O & XO \\
location-GPE & geographical entity such as cities, states, countries, and political entities \\
person-politician & politicians such as presidents, senators, and other government officials \\
organization-education & education institutions such as schools, colleges, and universities \\
\bottomrule
\end{tabular}
\caption{Extract of long labels for FewNERD.}
\label{table:long_fewnerd_labels}
\end{table}

\section{Validation Experiment with Sparse Latent Typing}
\label{sec:appendix_validation_experiment_sparse_latent_typing}

\begin{figure*}
    \centering
    \includegraphics[width=\textwidth]{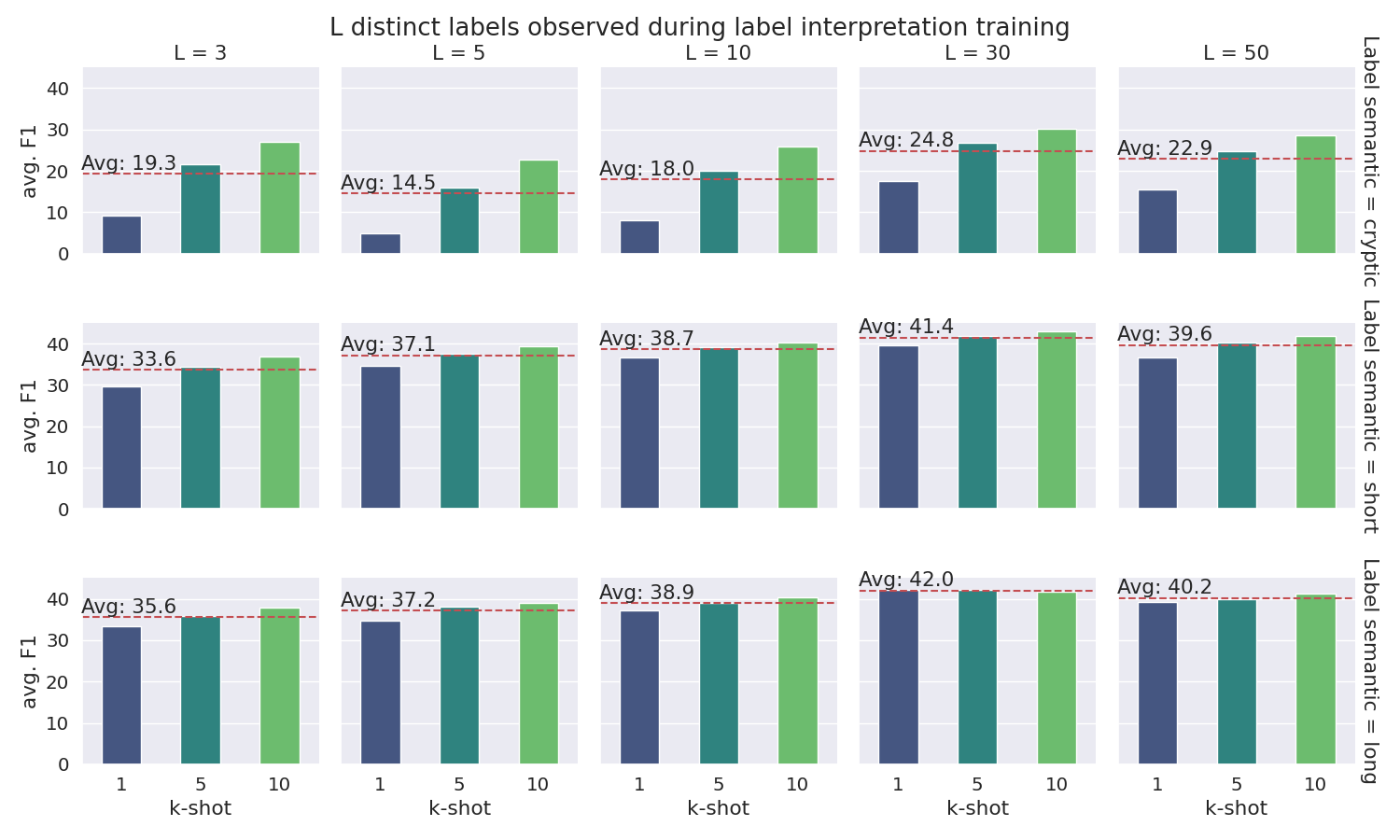}
    \caption{$K$-shot tagset extension on the 16 least occurring labels of FewNERD using the \texttt{sparse-latent-typing} encoder. We sweep over different numbers of distinct entity types and different semantic descriptions observed during label interpretation learning. We find that increasing both dimensions (more distinct types, extensive label verbalizations) contributes to an improved few-shot generalization.
    }
    \label{fig:appendix_motivation_graphic_sparse_latent_typing}
\end{figure*}

We perform our validation experiment on the recently released transformer using the sparse latent typing pre-training objective \citep{ren-etal-2022-language}. The experimental setup, including few-shot splits, is identical to the one in~\Cref{sec:validation_experiment}. The results are depicted in~\Cref{fig:appendix_motivation_graphic_sparse_latent_typing}.

Similar to the results in~\Cref{sec:validation_experiment}, we observe a better few-shot generalization with more distinct types and increased expressiveness of label verbalizations. However, the overall performance is higher using the encoder with sparse latent typing pre-training, a dedicated pre-training objective for keyword extraction from sentences. Further, we observe a slight decrease in performance as soon as $L$>30. This finding indicates that LitSet is transferable to entity-specific pre-trained models.

\section{WikiData labels} \label{sec:appendix_prefinetuning_labels}

Given all entity mentions from the entity linking dataset, we source various information from WikiData in natural language and annotate those entities with it. In the following, we present the selected attributes along with their respective definitions, which will serve as our labels:

\begin{enumerate}
    \item \texttt{x instance-of y}: Entity \texttt{x} is a particular example and instance of class \texttt{y}. For example, entity K2 is an instance of a mountain.
    \item \texttt{y subclass-of z}: Instance \texttt{y} is a subclass (subset) of class \texttt{z}. For example, instance class volcano is a subclass of a mountain.
    \item \texttt{description}: A short phrase designed to disambiguate items with the same or similar labels.
\end{enumerate}

 We note that the \texttt{instance-of} and \texttt{subclass-of} categories commonly encompass multiple tags rather than being limited to a single tag, as demonstrated in the example in \Cref{fig:zelda_annotation}. We filter out WikiData-related entities such as information or distribution pages because they do not contain any entity-related information.

\section{Hyperparameters}
\label{sec:lit_hyperparameters}

This section gives a detailed overview of the hyperparameters used throughout all experiments. For our baselines in experiments~\Cref{sec:validation_experiment,sec:extension_in_domain,sec:extension_out_domain,sec:cross_lingual} and~\Cref{sec:appendix_validation_experiment_sparse_latent_typing} we take the same hyperparameters as in \citep{ma-etal-2022-label} for label interpretation learning. An overview is listed in~\Cref{tab:lit_hyperparameters}.

\begin{table}[ht]
\centering
\begin{tabular}{ll}
\toprule
Argument & Value \\
\midrule
Learning rate & $1e^{-5}$ \\
Optimizer & AdamW \\
Scheduler & Linear warm-up (10\%) \\
Training epochs & 3 \\
Training batch size & 16 \\
Evaluation batch size & 16 \\
\bottomrule
\end{tabular}
\caption{We use \texttt{S-BERT} (all-mpnet-base-v2) and \texttt{SLT} (sparse latent typing) as the label encoder. \methodname{} transfers to other transformers and outperforms baselines in INTRA settings while remaining competitive in INTER settings with in-domain trained models.}
\label{tab:lit_hyperparameters}
\end{table}

\begin{table*}[ht]
\centering
\small
\renewcommand{\arraystretch}{1.3}
\begin{tabular}{lp{2.5cm}p{3.0cm}cccr}
\toprule
Transformer & Tagset extension
on $\mathcal{D}^{FS}$ & Label interpretation learning on $\mathcal{D}^{LIT}$ & 1-shot & 5-shot & 10-shot & Average \\
\toprule
\multirow{4}{*}{\texttt{S-BERT}} & \multirow{2}{*}{{FewNERD$_{\textsc{Intra}}$}} 
 & \methodname{} & $\bm{27.6} \pm 4.1$ & $\bm{49.2} \pm 3.4$ & $\bm{54.7} \pm 4.8$ & $\bm{43.8}$ \\
 & & {FewNERD$_{\textsc{Intra}}$} & $10.7 \pm 7.4$ & $37.8 \pm 9.8$ & $49.1 \pm 8.4$ & $32.5$ \\
\cline{2-7}
 & \multirow{2}{*}{FewNERD$_{\textsc{Inter}}$} & \methodname{} & $\bm{36.6} \pm 2.0$ & $\bm{44.3} \pm 2.0$ & $47.7 \pm 2.1$ & $\bm{42.9}$ \\
& & FewNERD$_{\textsc{Inter}}$ &  $23.4 \pm 2.4$ & $42.3 \pm 3.8$ & $\bm{48.5} \pm 3.1$ & $38.1$ \\
\midrule
\multirow{4}{*}{\texttt{SLT}} & \multirow{2}{*}{FewNERD$_{\textsc{INTRA}}$} & \methodname{} &  $\bm{27.2} \pm 5.8$ & $\bm{51.8} \pm 4.9$ & $\bm{57.2} \pm 5.4$ & $\bm{45.4}$ \\
& & FewNERD$_{\textsc{Intra}}$ & $6.2 \pm 4.9$ & $15.6 \pm 4.7$ & $21.9 \pm 4.9$ & $14.6$ \\
\cline{2-7}
& \multirow{2}{*}{FewNERD$_{\textsc{Inter}}$} & \methodname{} & $38.6 \pm 3.6$ & $49.4 \pm 2.5$ & $52.4 \pm 2.3$ & $46.8$ \\
& & FewNERD$_{\textsc{Inter}}$ & $\bm{40.3} \pm 4.1$ & $\bm{52.0} \pm 3.0$ & $\bm{54.9} \pm 2.24$ & $\bm{49.1}$\\
\bottomrule
\end{tabular}
\caption{We use \texttt{S-BERT} (all-mpnet-base-v2) and \texttt{SLT} (sparse latent typing) as the label encoder. \methodname{} transfers to other transformers and outperforms baselines in INTRA settings while remaining competitive in INTER settings with in-domain trained models.}
\label{tab:ablation_transfer_transformers}
\end{table*}

For \methodname{} in the respective sections, we use a lower learning rate of $1e^{-6}$, which achieved the lowest validation loss on a 5\% hold-out split of \methodname{}.

For few-shot fine-tuning, we use a slightly higher learning rate of $5e^{-6}$ for \methodname{} while the learning rate for the baselines remains at $1e^{-5}$. We use a maximum of 100 training epochs with early stopping after 5 iterations with no improvements on the training loss. We do not use any validation splits in few-shot fine-tuning for model selection.

All previous hyperparameters are identical for LEAR and BINDER (cf.~\Cref{sec:advanced_architectures}), except that we use the recommended learning rate of $3e^{-5}$ for BINDER and early stopping for label interpretation learning (after one epoch with no improvements on the training loss).

\section{Using Different Transformers as Label Encoder} \label{sec:ablation_sbert}

In this experiment, we investigate whether the \texttt{all-mpnet-base-v2} sentence transformer \cite{reimers-gurevych-2019-sentence} and the \texttt{sparse-latent-typing} transformer \citep{ren-etal-2022-language} can effectively help to understand label semantics better. Sentence transformers have been trained on a similarity objective, making them intriguing for our model to act as an enhanced label encoder. Sparse latent typing is a pre-training objective designed for extracting keywords from sentences. We present results in~\Cref{tab:ablation_transfer_transformers}.

We observe that using \texttt{all-mpnet-base-v2} performs generally worse than plain \texttt{bert-base-uncased}. However, we also observe that using \methodname{} yields better few-shot generalization in both INTRA and INTER settings and thus confirms that our main findings are transferable to other label encoders. When using \texttt{SLT} encoder, we outperform the baseline by large margins in the INTRA settings but fall slightly short in INTER settings.

\section{The Impact of Negative Examples} \label{sec:ablation_negative_sampling}

\begin{table*}[ht]
\centering
\small
\renewcommand{\arraystretch}{1.3}
\begin{tabular}{p{3.3cm}p{4.5cm}cccr}
\toprule
Evaluation data $\mathcal{D}^{FS}$ for tagset extension from:  & Label interpretation learning data $\mathcal{D}^{LIT}$ from: & 1-shot & 5-shot & 10-shot & Average \\
& (/w \# max. negative labels per batch) & & & \\
\toprule
\multirow{3}{*}{FewNERD$_{\textsc{Intra}}$} & \methodname{} (0) &$\bm{20.1} \pm 5.0$ & $\bm{47.7} \pm 6.0$ & $\bm{54.1} \pm 5.9$ & $\bm{40.6}$ \\
& \methodname{} (64) & $\bm{20.1} \pm 4.8$ & $47.5 \pm 5.0$ & $53.2 \pm 6.6$ & $40.3$ \\
& \methodname{} (128) & $18.9 \pm 4.9$ & $46.4 \pm 3.9$ & $52.7 \pm 5.9$ &  $39.3$ \\
\midrule
\multirow{3}{*}{FewNERD$_{\textsc{Inter}}$} &  \methodname{} (0) &$\bm{36.1} \pm 4.7$ & $47.2 \pm 3.0$ & $50.4 \pm 2.4$ & $\bm{44.6}$ \\
& \methodname{} (64) & $35.2 \pm 4.1$ & $\bm{47.4} \pm 2.6$ & $\bm{50.5} \pm 2.4$ & $44.4$ \\
& \methodname{} (128) & $34.7 \pm 3.3$ & $47.3 \pm 2.7$ & $50.4 \pm 2.3$ & $44.1$ \\
\bottomrule
\end{tabular}
\caption{The few-shot generalization of \methodname{} does not improve with a fixed number of labels per batch (we sample additional labels for loss calculation until, e.g., 64 labels are present). We find that the best training setup only uses the labels in the current batch.}
\label{table:negatives}
\end{table*}

In this experiment, we investigate the impact of integrating negative labels $\mathcal{L}^-$ in each batch. To do so, we additionally sample negative labels from $\mathcal{L} \setminus \mathcal{L}_b$ until the desired number of labels is reached and include them for loss calculation. Including negative types could potentially lead to a better generalization in few-shot settings due to the increased signal during loss calculation. We show results in \Cref{table:negatives}. We observe that including more labels in each batch harms the performance. While prior work \citep{epure-hennequin-2022-probing,wang-etal-2022-formulating} has shown that this idea is beneficial in few-shot settings, we find that \methodname{} works best when only using the labels present in the batch for loss calculation. Since we randomly sample additional labels, it is possible, if not likely, to sample similar labels that are not true negatives and thus not advantageous when using cross-entropy loss.

\section{Annotation Noise in ZELDA} \label{sec:annotation_noise}

In some cases, ZELDA is not consistently annotated, which may affect the few-shot fine-tuning performance for settings with very low $k$. \Cref{table:annotation_noise} shows such an example. We find unique entities, such as proteins, that are not consistently annotated to verify this assumption qualitatively. These inconsistencies may cause a worse entity detection ability with \methodname{} than training on consistently annotated datasets. While we show that entity linking benchmarks can be used to obtain a strong label understanding prior, improving the annotation quality or generating a designated label interpretation training dataset remains for future work.

\begin{table}[ht]
\centering
\small
\renewcommand{\arraystretch}{1.3}
\begin{tabular}{p{1.35cm}p{5.3cm}}
\toprule
\multicolumn{2}{c}{Annotation noise in ZELDA} \\
\toprule
annotated & {[\dots]} which in turn creates the compound \colorbox{pink}{oxyhemoglobin | \textbf{protein}}. \\
missing annotation & {[\dots]} whereas in \colorbox{lightgray}{oxyhemoglobin | \textbf{O}} it is a high spin complex. \\
\midrule
annotated &GSTK1 promotes \colorbox{pink}{adiponectin | \textbf{protein}} multimerization \\ 
missing annotation & {[\dots]} ER stress induced \colorbox{lightgray}{adiponectin | \textbf{O}} downregulation {[\dots]} \\
\bottomrule
\end{tabular}
\caption{Annotations in the entity linking benchmark may be inconsistent, causing the 1-shot drops on JNLPBA. Since JNLPBA is annotated by humans, it is expected that all sentences are annotated consistently.}
\label{table:annotation_noise}
\end{table}

\end{document}